\documentclass[journal]{IEEEtran}
\usepackage[latin1]{inputenc}
\usepackage[dvipsnames]{xcolor}
\usepackage{array}
\usepackage{tabularx}
\usepackage{float}
\usepackage{multicol}
\usepackage{booktabs}%

\newcommand{\tabitem}{~~\llap{\textbullet}~~}
\usepackage{multirow}
\usepackage{siunitx}
\usepackage{amsmath}

\ifCLASSOPTIONcompsoc
  \usepackage[nocompress]{cite}
\else
  \usepackage{cite}
\fi
\usepackage{multirow}

  \usepackage[pdftex]{graphicx}
\ifCLASSOPTIONcompsoc \usepackage[caption=false,font=footnotesize,labelfont=sf,textfont=sf]{subfig}
\else \usepackage[caption=false,font=footnotesize]{subfig}
\fi
\begin{document}
%
\title{DASPS: A Database for Anxious States based
on a Psychological Stimulation}
%
%
%

\author{Asma~Baghdadi,~\IEEEmembership{~Member,~IEEE,}
		Yassine~Aribi,~\IEEEmembership{Senior~Member,~IEEE,}
        Rahma~Fourati,~\IEEEmembership{~Member,~IEEE,}
        Najla~Halouani,
        Patrick~Siarry,~\IEEEmembership{Senior~Member,~IEEE}
        and~Adel M.~Alimi,~\IEEEmembership{Senior~Member,~IEEE}
\IEEEcompsocitemizethanks{\IEEEcompsocthanksitem A. Baghdadi, Y. Aribi, R. Fourati and A. M. Alimi are with the Research Groups in Intelligent Machines, National Engineering School of Sfax (ENIS), University of Sfax,
BP 1173, 3038 Sfax, Tunisia\protect\\
E-mail: \{asma.baghdadi, yassine.aribi, rahma.fourati, adel.alimi\}@ieee.org
\IEEEcompsocthanksitem Najla Halouani is with the Hedi Chaker hospital, Sfax, Tunisia.\protect\\
E-mail:najla.halouani@yahoo.fr
\IEEEcompsocthanksitem Patrick Siarry is with the LERISS, Universit\'e de Paris 12, 94010 Cr\'eteil, France.\protect\\
E-mail:siarry@univ-paris12.fr}}

\maketitle
\begin{abstract}
Anxiety affects human capabilities and behavior as much as it affects productivity and quality of life. It can be considered as the main cause of depression and suicide. Anxious states are
detectable by specialists due to their acquired cognition and skills, humans interpret the interlocutor's tone of speech, gesture, facial
expressions and recognize their mental state. There is a need for non-invasive reliable techniques that performs the complex task of
anxiety detection. In this paper, we present DASPS database containing recorded Electroencephalogram (EEG) signals of 23 participants during anxiety elicitation by means of face-to-face psychological stimuli. EEG signals were captured with Emotiv Epoc headset as it's a wireless wearable low-cost equipment. In our study, we investigate the impact of different parameters, notably: trial duration, feature type, feature combination and anxiety levels number. Our findings showed that anxiety is well elicited in 1 second. For instance, stacked sparse autoencoder with different type of features achieves 83.50\% and 74.60\% for 2 and 4 anxiety levels detection, respectively.
The presented results prove the benefits of the use of a low-cost EEG headset instead of medical non-wireless devices and create a starting point for new researches in the field of anxiety detection.

\end{abstract}




%
\IEEEpeerreviewmaketitle
\section{Introduction}
\label{sec:introduction}

\IEEEPARstart{A} {nxiety} is a mental health issue that has physical consequences on our bodies. However, it can affect the immune system, and unfortunately, there is evidence that too much anxiety can actually weaken the immune system dramatically \cite{adam2018}. Anxiety is essentially a long term stress, in such a way the stress hormone is liberated by our bodies in huge quantities which correlates with body performance degradation. This invisible disability can greatly affect academic performance as well. Anxiety impacts memory capacities, leading to difficulties in learning and retraining information. 

An anxious individual performs less efficiently, which significantly affects his capabilities. One in eight children suffers from anxiety disorders according to the Anxiety Disorders Association of America reports \cite{adaa}. Nevertheless, it presents a risk for poor performance, diminished learning and social/behavioral problems in school. Since anxiety disorders in children are difficult to identify, it is an imperative task to learn how to detect them in early stages in order to help them. It may manifest by signs such as increased inflexibility, overreactivity and emotional intensity.

Besides, effective anxiety and stress management can help balance stress in your life, while keeping high productivity and enjoying life. The intention is to find harmony between work, relationships and self-awareness, and to learn how to deal with anxiety states to confront challenges. But anxiety management is not one-size-fits-all, so we need to detect when anxiety is present, how it manifests in our bodies and how our neurological system reacts to such situations.
 
 Recently, Dr. Shanker introduced the Self-Reg Framework which helps individuals understand and interact with others by considering self-regulation from the biological, emotional, cognitive, social, and pro-social domains using The Shanker Method: Reevaluate the behavior.Identify the stressors, Minimize the stress, Enhance stress awareness and Develop personalized strategies to promote resilience and restoration. Hence, anxiety and stress management is one of numerous fields that requires the need for automated anxiety detection.  
 
Anxiety detection is an underlying part of affect recognition. 
Another area that could significantly benefit from progress in affect recognition is the video game industry. New trends of therapeutic environments for rehabilitation of patients with serious mental disorders, implement an affect detection algorithms. Biofeedback systems can help children, adolescent and adults control and manage their levels of anxiety, and facilitate real life challenges. 

For our research we treated anxiety as a temporary state as such, we selected healthy patients and we used the recitation of real life situations to stimulate anxiety within them.\\    
Anxiety mainly arises due to three factors, namely external, internal and interpersonal. Table \ref{anxietytrig} shows anxiety categories and their stimuli from real life situations. To select the situations with the highest anxiety levels, a survey was carried out and diffused for all volunteers who wanted to participate in our experiment. Based on the conducted survey, we selected 6 situations where participants experienced the highest anxiety levels as portrayed hereinafter: Loss (68\%), Family issues (64\%), financial issues (54\%), Deadline (46\%), Witnessing deadly accident (45\%) and Mistreating (40\%). 
\begin{table}[!t]
\scriptsize
\renewcommand{\arraystretch}{1.3}
  \centering
   \caption{Anxiety triggers categories and stimuli}\label{anxietytrig}
\begin{tabular}{lll}
\toprule
  Category & Stimuli \\
\midrule
  External & \tabitem Witnessing a deadly accident \\
           & \tabitem Familial instability / Financial instability / Maltreatment / Abuse \\
           & \tabitem  Deadlines / Insecurity / Routine \\
           & \\
  Interpersonal & \tabitem Relationship with the supervisor / manager \\
          
           & \tabitem Lack of confidence towards spouse \\
           & \tabitem Being in an embarrassing situation\\
           & \\
  Internal & \tabitem Fear of getting cheated on / Fear of losing someone close  \\
	       
           & \tabitem Fear of children's failure / Feeling guilty permanently \\
           
           & \tabitem Recalling a bad memory \\
           & \tabitem Health (Fear of getting sick 
            and missing on an important event)\\
           & \tabitem Health (Fear of being diagnosed with a serious illness)
 \\
\bottomrule
\end{tabular}
\end{table}

This paper is organized into 8 sections. In section 2, we present an overview of the realized works for anxiety detection from EEG signals. In section 3, we detail the
implemented experimental protocol and the analysis of the collected data in terms of variability and coherence. The followed steps for data recording and preprocessing as well as the general architecture of the proposed system are presented in section 4. A variety of features are presented in Section 5. Three classifiers are described in Section 6. An analysis and discussion of the obtained results are carried out in section 7. Finally, the last section summarizes our paper and outlines future work.
\section{Literature overview}
\begin{table*}[!t]

\renewcommand{\arraystretch}{1.3}
\caption{Previous works on EEG-based Anxiety Detection}
\label{existingworks_table}
\centering
\begin{tabular}{c|c|c|c|p{3cm}|p{2cm}|c}
\hline \hline
\bfseries Reference & \bfseries Stimulus & \bfseries \#Participants& \bfseries \#Channels & \bfseries Method description & \bfseries Affective states & \bfseries Accuracy (${\%}$)\\ 
\hline
\cite{fourati2018unsupervised}&Audio-Visual&32&32&
ESN with band power features&Stress and Calm&76.15\\ \hline
\cite{garcia2017symbolic}&Audio-Visual&32&32&SVM with Entropy features&Stress and Calm&81.31\\
\hline
\cite{katsigiannis2018dreamer}& Audio\-Visual & 23 & 14 & PSD with SVM& Valence, Arousal and Dominance & 62.49\\
\hline
\cite{giannakakis2015detection}&Audio-Visual&18&32&Asymmetry Index, Coherence, Brain Load Index and Spectral Centroid Frequency&Stress and Relax&-----\\
\hline
\cite{vanitha2016real}&Mathematical tasks&6&14&Hilbert-Huang Transform with SVM&Neutral, Stress-low, Stress-medium and Stress-high&89.07\\
\hline
\cite{patil2017method}&-----&13&8&k-means clustering with stress indice &Stress and Relax&----\\
\hline
\cite{lim2015analysis}&Stroop Color Word test&25&1&ANN, k-NN, LDA with DCT coefficient &Non-stressed and Stressed&72.00\\
\hline
\cite{khosrowabadi2011brain}&Examination period&26&8&k-NN and SVM with Higuchi FD, GM and MSCE &Stress and Stress-free&90.00\\
\hline \hline
\end{tabular}
\end{table*}
Based on biometrics, many researches were conducted for the recognition of persons' identity, mental and physical health \cite{aribi2015automated}  \cite{aribi2013intelligent} \cite{aribi2013automatic} \cite{aribi2013system} \cite{aribi2012analysis} \cite{aribi2014automated}. \\
Researches conducted for anxiety/stress detection based on EEG signals analysis are few compared to those done for emotion recognition surveyed in  \cite{baghdadi2016survey} \cite{alarcao2017emotions}. Most of the proposed works for EEG-based emotion recognition as in \cite{fourati2017optimized} \cite{fourati2018unsupervised} were validated using DEAP datatset \cite{koelstra2012deap}. It was recorded using Biosemi Active 2 headset with 32 channels. 32 particpants were watching 40 videos of one-minute duration.
In their work, Giorgos \emph{et al.} \cite{giannakakis2015detection} extracted two subdatasets of trials from DEAP dataset \cite{koelstra2012deap} according to predefined conditions for two emotional states: stress and calm. The idea is to define thresholds for valence and arousal and extract only trials which respect this condition. Consequently, the previous step leads to a subset of 18 subjects conform to the adequate norm. The authors extracted spectral, temporal and non linear EEG features to represent the investigated states.
%
%
Otherwise, some researchers opt to conduct a suitable experimentation to collect their own EEG signals. In the work of Vanita \emph{et al.} \cite{vanitha2016real}, the authors looked at students' stress levels and defined their own experimental protocol to record EEG signals during stress elicitation session. Thenceforth, data was preprocessed for noise and ocular artifact removal. Features were extracted by the mean of a time-frequency analysis and classification was performed by an hierarchical Support Vector Machines
that gave an accuracy of 89.07\%. To investigate the realtime issue, Lahane \emph{et al.} \cite{lahan2016} proposed an EEG-based stress detection system. They employed an android application to gather EEG data. As feature, the Relative Energy Ratio (RER) was calculated for each frequency band.

Single channel EEG signal was recorded from 25 students from the Sunway University for a Stress Detection System proposed by \cite{lim2015analysis}. Using the NeuroSky Mindwave headset, the data was collected and stored for further analysis. Students' stress was elicited for 60 seconds by Stroop color word test preceded by 30 seconds of one-screen instruction reading. EEG signal's high-frequency components comprising noise and artifacts were discarded, and only low-frequency components obtained after a Discrete Cosine Transform (DCT) were processed to the classification. 
Based on an interview with the subjects who reported that the instruction reading was the most stressful part of the experiment, it was concluded that only the first 30 seconds of the recorded data were preprocessed and processed for stress classification. Results show that k-NN which reaches 72\% outperforms LDA(60\%) and ANN(44\%) in classifying stress. 

Khosrowabadi \emph{et al.} \cite{khosrowabadi2011brain}, discerned that the examination period is the most stressful for students. Based on this fact, they conducted their experiment during and after the examination period. Collecting EEG signals from 26 students (15 during examination period and 11 two weeks after). Data were preprocessed to noise removal with an elliptic band-pass filter (2-32 Hz). Three different features were investigated in this work: Higuchi's Fractal Dimension (HFD), Gaussian mixtures of EEG spectrogram and Magnitude Square Coherence Estimation (MSCE). The classification step was handled using k-NN and SVM classifiers. As a result, MSCE gives the best accuracy up to 90\% in classifying chronic mental stress. 
To detect stress in healthy subjects, Norizam \emph{et al.} \cite{sulaiman2011novel} used in their study a k-NN classifier and some parameters such as Shannon Entropy (SE), Relative Spectral Centroid (RSC) and Energy Ratio (ER). Their study employed 185 EEG data from different experiments. 
%
%

According to Table \ref{existingworks_table}, most of the previous works rely on audio-visual stimulus from the international IAPS, IADS databases \cite{oude2007eeg}. Whereas, others used arithmetic tasks as practiced in \cite{jun2016eeg} where the level of stress is supposed to increase when the hardness level of tasks is increased. In our opinion, mathematical tasks can not be realized by any person. Thus, the experience is limited to specific participants. Let's recall that anxiety is felt by any person and our aim is to detect it in all people without any restriction. The definition of anxious states differs from a study to another. In our work, we will keep the term anxiety and present its levels as recommended by our therapist.
For the classification, k-NN and SVM are popular among all works. Therefore, we follow these works and use them for the detection of anxiety levels.  To add, we also use Stacked Sparse 
AutoEncoder (SSAE) for the classification step.

In this context, we rise the challenge of proposing a new database \footnote{The database and Matlab scripts for data segmentation considered in this article can be downloaded on the following
site: http://www.regim.org/publications/databases/dasps/} 
of EEG signals for anxiety levels detection. The innovation of our work does not reside only in making  public EEG data for affective computing community but also in the design of a psychological stimulation protocol providing comfortable conditions for participants thanks to the face-to-face interaction with the therapist and to the use of a wireless EEG cap with lesser channels, i. e. only 14 dry electrodes. 
\section{Experimental protocol}
\subsection{Protocol description}
We defined an experimental protocol that meets our needs. After a discussion with a psychotherapist, she recommended the use of exposure therapy. The latter is a form of the well known Cognitive Behavioral Therapy (CBT). Exposure therapy involves starting with items and situations that cause anxiety, but one that you feel able to tolerate \cite{eraldi2007exposition}. 
There are different forms of exposure, such as Imaginal Exposure, virtual reality exposure, and in-vivo exposure.
In our experimental protocol we used the Flooding as in-vivo exposure therapy \cite{eftekhari2006you}, actual exposure to the feared stimulus. A patient is confronted with a situation in which the stimulus that provoked the original trauma is present. Regarding the conditions of our experimentation, we used Flooding since it is quick and usually effective.\\
The fixed protocol is as follows. Each participant is asked to sign a consent before starting the experiment. The anxiety level is calculated before stimulation according to the Hamilton
Anxiety Rating Scale (HAM-A) in order to measure the
severity of participants' anxiety. This tool provides 14 items, each one contains a number of symptoms that can be rated on a scale of zero to four. 

Our psychotherapist inquires the participant about the degree of severity of each symptom and its rate on the scale, with four being the most severe. This acquired data is used to compute an overarching score that indicates a person's anxiety severity \cite{mcevoy2015physiologic}. After which, the participant is prepared to start the experiment, with closed eyes and minimizing gesture and speech. The psychotherapist starts by reciting the first situation and helps the subject to imagine it. This phase is divided into two stages: recitation by the psychotherapist for the first 15 sec and Recall by the subject for the last 15 sec. 

\begin{figure}[!t]
  \centering
  \includegraphics[width=3.5in]{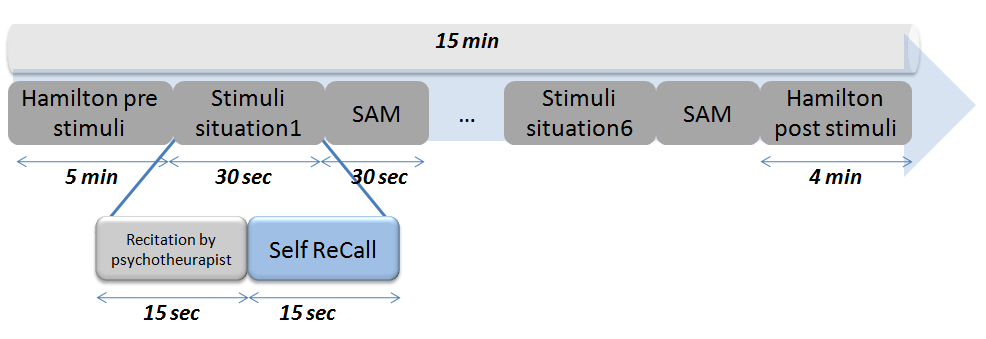}
  \caption{The experimental protocol of anxiety stimulation}\label{protocol}
\end{figure}

When time is over, the subject is asked to rate how he felt during stimulation using the Self Assessment Manikin (SAM). It has two rows for rating: Valence ranging from negative to
positive and Arousal ranging from calm to excited. Each row contains nine items for rating. In order to evaluate the current emotion, each volunteer has to tick items that are suitable for emotion on only two dimensions (Arousal, valence). This trial is repeated until the sixth situation. At the end of the experiment, some items from HAM-A are re-evaluated by the psychotherapist to adjust the participant's anxiety level. All steps of the protocol of stimulation is presented in Fig. \ref{protocol}.
\subsection{Data analysis}
\begin{figure}[!t]
  \centering
  \includegraphics[width=2.5in]{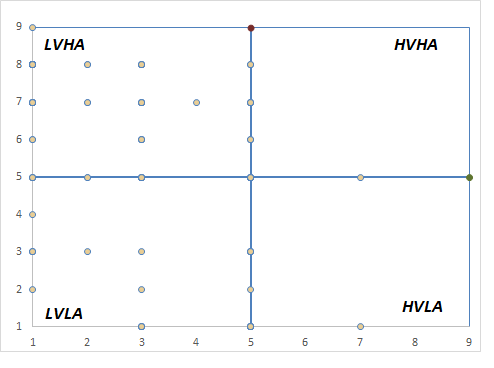}
  \caption{Presentation of participant rating in two-dimensional space}
  \label{plan}
\end{figure}
Before the preprocessing phase, the data were evaluated to eliminate those with large difference between expected and real rating.
In \cite{russell1977evidence} \cite{russell1980circumplex}, Russell defines anxiety as: Low Valence and High Arousal. As a matter of fact, trials having this condition and belonging to LVHA quadrant are the main focus of our work as shown in Fig. \ref{plan}.
To analyze data across all participants, we opt to measure the relative variability by computing the Coefficient of Variation (CV) for all participants' ratings for all stimuli situations. 

CV is the ratio of the standard deviation to the mean (average).
If the calculated CV is equal to zero, it indicates no variability whereas higher CVs indicate more variability.
The mean CV between the participants' assessments was 0.58 for valence and 0.42 for arousal, which can be considered as low variability. Note that this value is higher than expected. In most cases, this variability is due to the lack of comprehension of SAM scales leading to no objective rating.
The mean rating across all study participants for each stimuli case in terms of valence and arousal are shown in Table \ref{evaltab3}.

For each situation, the participant rating can be presented in 2D plan corresponding to the valence and arousal values. This plan can be divided into four quadrants according to the possible combinations of valence and arousal scales. The four quadrants as shown in Fig. \ref{plan} are:
Low Valence and Low Arousal (LVLA), High Valence and Low Arousal (HVLA), Low Valence and High Arousal (LVHA), and High Valence and High Arousal (HVHA).
A summary of the subjective classification into the four Valence-Arousal quadrants from participants' ratings is presented in Table \ref{evaltab}. 

As shown in Fig. \ref{plan}, samples are focused on LVHA and LVLA quadrants, which proves that the employed situations successfully worked in eliciting anxiety with most participants. Table \ref{evaltab2} presents the number of participants classified by anxiety levels based on HAM score before and after the experiment.The number of participants with severe anxiety increased from 7 to 13 denoting the impact of elicitation of participants' anxiety level.
\begin{table}[!t]
\renewcommand{\arraystretch}{1.3}
  \centering
  \caption{Participants Number in each quadrant according to SAM ratings}
  \label{evaltab}
  \begin{tabular}{c|c|c|c|c}
\hline \hline
   Stimulus  & LVLA & HVLA & LVHA & HVHA \\
     \hline 
    Situation 1 & 7 & 0 & 16 & 0 \\ \hline
    Situation 2 & 12 & 0 & 11 & 0 \\ \hline
    Situation 3 & 9 & 0 &  14 &0 \\ \hline
    Situation 4 & 14 & 0 & 7 & 0 \\ \hline
    Situation 5 & 8 & 0 & 15 & 0 \\ \hline
    Situation 6 & 7 & 0 & 16 & 0 \\
    \hline \hline
  \end{tabular}
\end{table}

\begin{table}[!t]
\renewcommand{\arraystretch}{1.3}
  \centering
  \caption{Participants anxiety levels according to Hamilton scores}
  \label{evaltab2}
  \begin{tabular}{c|c|c|c|c}
\hline \hline
    Anxiety level & Normal & Light & Moderate & Severe \\
    \hline
    Before Experiment & 4 & 6 & 6 & 7 \\ \hline
    After Experiment & 2 & 5 & 3 & 13 \\
    \hline \hline
  \end{tabular}
\end{table}

\begin{table}[!t]
\renewcommand{\arraystretch}{1.3}
  \centering
  \caption{Mean rating and Standard Deviation across all participants for each situation}
  \begin{tabular}{c|c|c}
\hline \hline
   Stimulus  & Valence & Arousal \\
     \hline
    Situation 1 & $2.13 \pm 1.68$ & $6.13\pm2.63$ \\ \hline
    Situation 2 & $3.43\pm1.44$ &$ 5.13\pm2.68$ \\ \hline
    Situation 3 & $1.86\pm1.25$ & $6.04\pm1.69$ \\ \hline
    Situation 4 & $3.86\pm1.79$ & $4.30\pm2.47$ \\ \hline
    Situation 5 & $3.30\pm1.63$ & $5.95\pm2.24$ \\ \hline
    Situation 6 & $2.26\pm1.54$ & $6.30\pm2.18$ \\
    \hline
    Mean CV & 0.58 & 0.42 \\
    \hline \hline
  \end{tabular}
  \label{evaltab3}
\end{table}

\begin{figure*}[!t]
  \centering
  \includegraphics[width=6in]{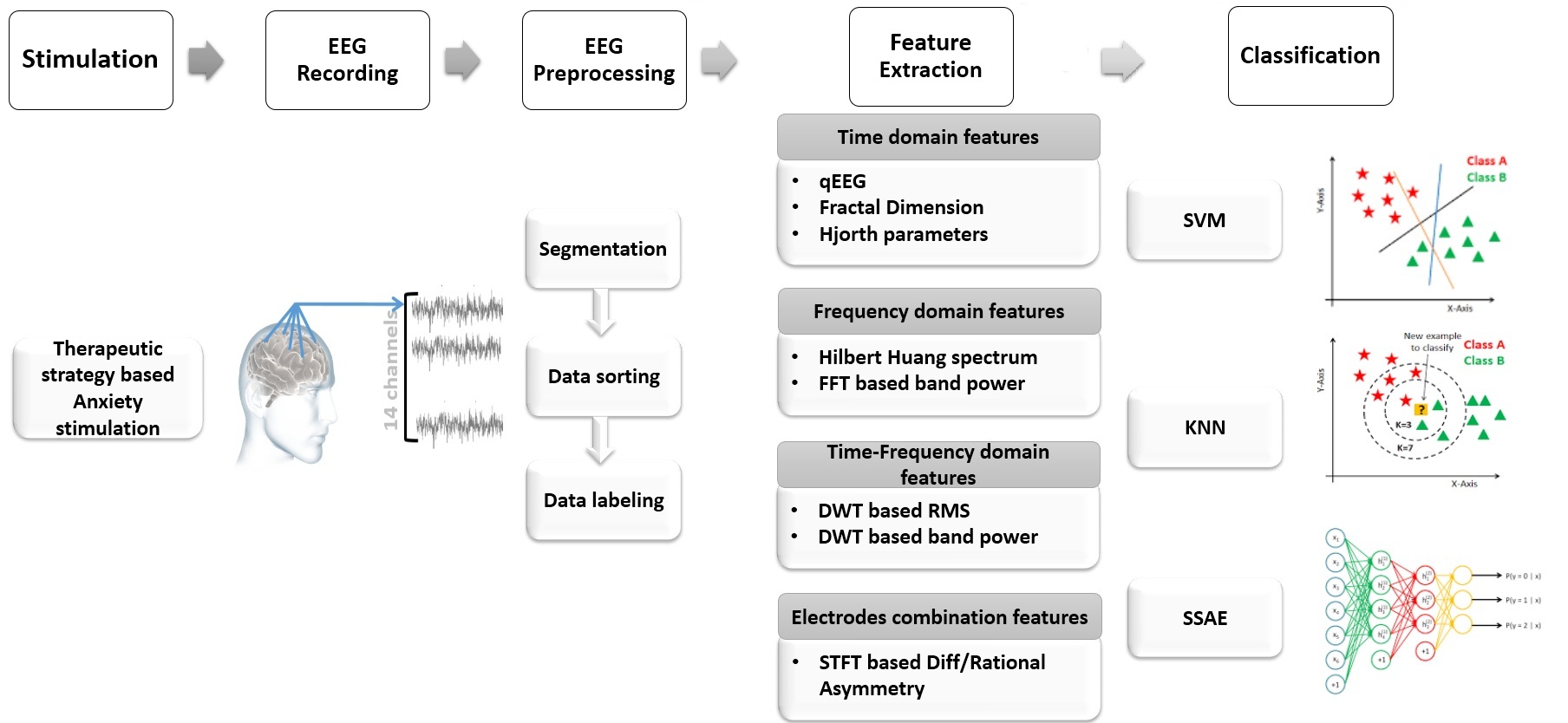}
  \caption{Architecture of the proposed system}
  \label{arch}
\end{figure*}
\section{Materials and Methods}
This work covers all stages needed to create a robust EEG-based anxiety detection system, starting from the elaboration of an anxiety stimulation experimental protocol to the classification of anxiety levels. The general architecture of the proposed system is depicted in Fig. \ref{arch}.
\subsection{EEG recording}
\begin{figure}[!t]
  \centering
  \includegraphics[width=1.5in]{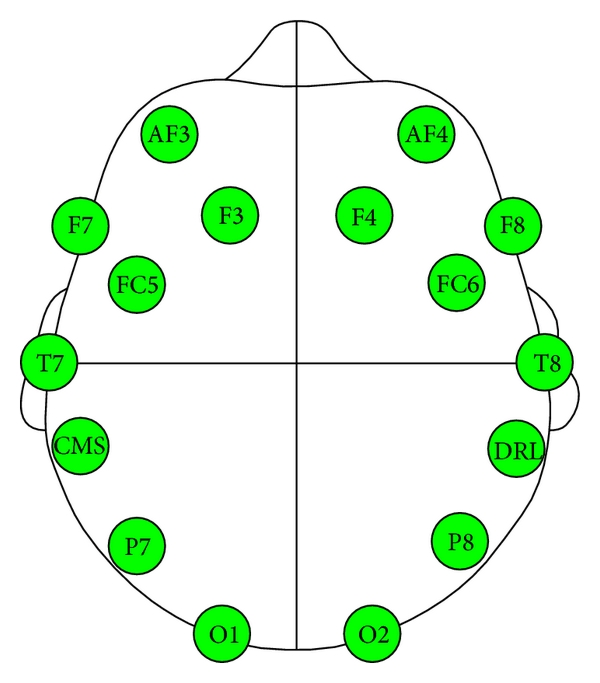}  \caption{Emotiv EPOC electrodes placement}\label{electrode}
\end{figure}
The experiment was performed on 23 healthy subjects not suffering  from psychological diseases. 13 women and 10 men with an average  age of 30 years old. The purpose was clearly explained to each  participant before starting the experiment. Items of the Hamilton
test are highlighted to avoid misunderstanding each question.
The experiment was performed in an isolated environment to avoid distracting noises and to guarantee a subject's full concentration. The anxiety stimulation is accomplished by face-to-face psychological elicitation performed in a professional manner by our psychotherapist. 
EEG signals were recorded using a wireless EEG headset, the Emotiv EPOC 14 channels and 2 mastoids \cite{ekanayake2010p300} were placed according to the international 10-20 system. The electrodes were attached to the scalp at position AF3, F7,F3, FC5, T7, P7, O1, O2, P8, T8, FC6, F4, F8 and AF4 as shown in Fig. \ref{electrode}. he M1 mastoid sensor acts as a ground reference point to compare the voltage of all other sensors, while the M2 mastoid sensor is an indirect reference to minimize electrical interference from the outside.\cite{katsigiannis2018dreamer}.

Emotiv Epoc neuro headset \cite{ekanayake2010p300} is used in this work regarding it's easiness of use. It offers comfort to users and mainly it is a wireless equipment and don't require an intricate set up like with clinical EEG material. In addition, it shows efficiency while used for emotion recognition systems like proved by \cite{liu2011real} \cite{jatupaiboon2013real} \cite{jatupaiboon2013emotion} \cite{anh2012real} \cite{coan2003frontal} and more recently by \cite{katsigiannis2018dreamer} \cite{benitez2016use}.

The recording was performed through Emotiv Epoc Software for EEG raw data recording.
It allows us to view and save data for
all channels or just customised the ones we need. The produced raw data have ".Edf" extension that is convertible using matlab script to ".mat" for further processing.
The recording started before carrying out the first
situation and ended after finishing the sixth one.
Each recording took 6 min as depicted in Fig. \ref{protocol}, divided into 1 min by trial. The acquired EEG signals were processed at 128 Hz and impedance was kept as low as 7 k\si{\ohm}.
\subsection{EEG preprocessing}
In biomedical signal processing, the determination of noise and artifacts in the acquired signal is necessary to move to the feature extraction phase with a clean signal and achieve good classification results. Physiological artifacts are generated by a source different than the brain, such as electroculogram (EOG) artifacts under 4 Hz, muscle artifacts (EMG) with frequency exeeding 30 Hz, and heart rate ( electrocardiogram: EMG) of about 1.2 Hz. They can also be extra physiological, unrelated to the human body and are in the 50 Hz range. This may be caused by the environment or related to EEG acquisition parameters \cite{oude2007eeg} \cite{mcevoy2015physiologic} .

For the aim of denoising our set of signals, we have applied an EEGLab script serving to cut relevant sub-band of EEG signals, removing baseline and removing Ocular and Muscular artifacts. A 4-45 hz Finite impulse response (FIR) pass-band filter was applied to the raw data.
The Automatic Artifact removal for EEGLAB toolbox \cite{delorme2004eeglab} (AAR) was used to remove EOG and EMG artifacts.\\
 This toolbox implements several algorithms for EMG and EOG artifacts removal. 
 We used the implementation of BSSCCA Canonical Correlation Analysis (CCA) which projects the observed EEG data into maximally auto-correlated components\cite{de2006canonical}. We chose the criterion emg\_psd that considers the components whose average power ratio in the typical EEG and EMG bands is below certain threshold to be EMG-related .
  In order to estimate the power in the EEG and EMG bands, the default estimator used is a Hamming-windowed Welch periodogram with segment length equal to the analysis window length.\\
  By default, the toolbox uses a combination of iWASOBI which is an asymptotically optimal Blind Source Separation (BSS) algorithm for autoregressive (AR) sources\cite{tichavsky2006computationally}, and the criterion eog\_fd to automatically correct EOG artifacts in the EEG. Eog\_fd marks as artifactual the components with smaller fractal dimension. Conceptually, components with low fractal dimensions are those that are composed of few low-frequency components \cite{gomez2006automatic}. This is often the case of ocular activity and therefore it is a suitable criteria for detecting ocular (EOG) components.
  The AAR toolbox offers a graphical interface, but we recommend scripts that facilitate and automate the denoising of all signals.

As mentioned, the experiment lasts almost 6 minutes divided into 6 different situations. We are only interested by the first 30 seconds of each trial. 15 seconds of SAM are removed during this step. As a result, we have 6 trials of 30 sec each per participant. 
We recall that, as shown in the experiment protocol, after each 30 sec of stimulation the participant is asked to fill in the SAM survey to express in terms of excitement (Arousal) and feeling (Valence) his emotions during the stimulation. And after the end of the whole experiment, the participant is asked to identify the most anxious situation. We used this information to label all trials.

By applying this labeling step, we get: 156 'Normal' trials, 90 'Severe' trials, 10 trials 'Moderate' and 20 'Light' trials.
In order to increase the number of samples per participant, we follow previous work \cite{zheng2017identifying} by constructing two additional sub datasets for 5 seconds and 1 second trials extracted as sample from the main EEG signal.

\begin{figure}[!t]
  \centering
  \includegraphics[width=3.5in]{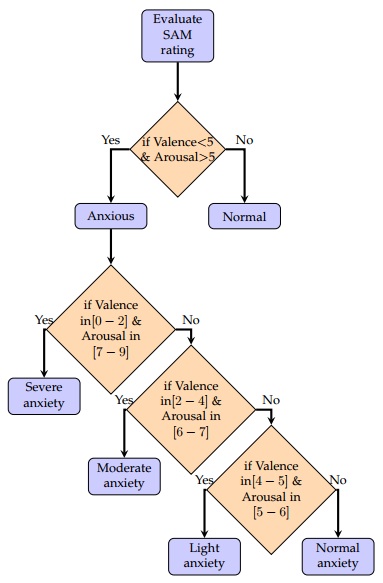}
  \caption{Flow Chart of the labeling process}\label{flow}
\end{figure}

The labeling process is depicted in the Flow Chart of Fig.\ref{flow}.
\section{Feature extraction}
A wide range of EEG features for emotion recognition have been investigated in the literature \cite{jenke2014feature}. Generally, we can classify these features into three main classes according to the domain, namely, time-domain features, frequency-domain features and time-frequency-domain features. Other features can be extracted from a combination of electrodes, we mention one of them in this section.
\subsection{Time Domain Features}
Time-domain features are results of an exploration of signal characteristics that differ between emotional states. Many approaches were employed in the researches to extract this type of features. In our work, we have extracted Hjorth features and FD features:
\subsubsection{Hjorth Features}
Hjorth parameters \cite{hjorth1970eeg} are: Activity, Mobility, and Complexity. The variance of a time series represents the activity parameter. The mobility parameter is represented by the mean frequency, or the standard deviation proportion of the power spectrum. Finally The complexity parameter represents the variation in frequency. Besides, it indicates the deviation of the slope.

Assume that
{$dx_{i}=x_{i+1} - x_{i'}(i=1,..,n-1), ddx = dx_{i+1} - d_{i'}(i=1,..,n-1)$}.\\
 The expressions of Hjorth parameters are:
\footnotesize{
   \begin{equation}\emph{Activity = $\frac{1}{n} \sum\limits_{i=1}^n x_{i}^2$}  ,  \emph{Mobility = $\frac{\frac{1}{n-1} \sum\limits_{i=1}^{n-1} dx_{i}^2}{\frac{1}{n} \sum\limits_{i=1}^n x_{i}^2 }$}\end{equation}
    \begin{equation}\emph{Complexity =  $ \sqrt{\frac{\frac{1}{n-2} \sum\limits_{i=1}^{n-2} ddx_{i}^2}{\frac{1}{n-1} \sum\limits_{i=1}^{n-1} dx_{i}^2}-\frac{\frac{1}{n-1} \sum\limits_{i=1}^{n-1} dx_{i}^2}{\frac{1}{n} \sum\limits_{i=1}^n x_{i}^2}} $}\end{equation}
}
\normalsize

Hjorth were used in many EEG studies such as in  \cite{ansari2007channel} \cite{hjorth1970eeg} \cite{horlings2008emotion}.
In our work, we calculated Hjorth parameters for all EEG channels, that produce a size Feature Vector of 42x1 for each trial.
\subsubsection{Fractal Dimension}
The Higuchi algorithm calculates fractal dimension
value of time-series data.
X (1), X (2),..., X (N) is a finite set of time
series samples.
Then, the newly constructed time series is defined
as follows:
\begin{equation}
X_{k}^{m}:X(m),X(m+k),...,X
\begin{pmatrix}
m+
\begin{bmatrix}
\frac{N-m}{k}
\end{bmatrix} .k
\end{pmatrix}
\end{equation}
\begin{equation}
  \left ( m=1,2,...,k \right )
\end{equation}

where m is the initial time and k is the interval time.
k sets of $ L_m(k) $ are calculated as follows:
\begin{equation}
\scriptsize
  L_{m}(k)=\frac{\begin{Bmatrix}
\begin{pmatrix}
\sum_{i=1}^{\begin{bmatrix}
\frac{N-m}{k}
\end{bmatrix}}\left | X(m+ik)-X(m+(i-1).k) \right |
\end{pmatrix}
 \frac{N-1}{\begin{bmatrix}
\frac{N-m}{k}
\end{bmatrix}.k}
\end{Bmatrix}}{k}
\end{equation}
where L(k) denotes the average value of $ L_m(k)$ , and a relationship exists as follows:
\begin{equation}
  \left \langle L(k)) \right \rangle\infty  k^{-D}
\end{equation}
Then, the fractal dimension can be obtained by
logarithmic plotting between different k and its
associated L(k).
\subsection{Frequency Domain Features}
\textbf{Band Power}
Power bands features are the most popular features in the context of EEG-based emotion recognition. The Definition of EEG frequency bands differs slightly between studies. Commonly,
they are defined as following: $\delta$ (1-4 Hz), $\theta$ (4-8 Hz), $\alpha$ (8-13 Hz), $\beta$ (13-32 Hz) and  $\Gamma$ (32-64 Hz).

The decomposition of the overall power in the EEG signal into individual bands is commonly achieved through Fourier transforms and related methods for spectral analysis \cite{saby2012utility} as stated in \cite{jenke2014feature}. Otherwise, short-time fourier transform (STFT) \cite{zheng2017identifying} is the most commonly used alternatives, or the estimation of power spectra density (PSD) using Welch's method \cite{koelstra2012deap}.

\subsection{Time-Frequency Domain Features}
\subsubsection{Hilbert-Huang Spectrum}
The Empirical Mode Decomposition (EMD) along with the Hilbert-Huang Spectrum (HHS)
are considered as a new way to extract necessary information from EEG signal since it defines
amplitude and instantaneous frequency for each sample \cite{panoulas2008hilbert}.
EMD decomposes the EEG signal into a set of Intrinsic Mode Function (IMF) through an automatic shifting process. Each IMF represents different frequency components of original signals. EMD acts as an adaptive high-pass filter.
It shifts out the fastest changing component first and as the level of IMF increases,
the oscillation of the latter becomes smoother. Each component is band-limited, which can reflect the characteristic of instantaneous frequency \cite{zhuang2017emotion}.

x(t) is then represented as a sum of IMFs and the residual \cite{panoulas2008hilbert}.
\begin{equation}\label{hht}
\footnotesize{
  x(t)=\sum_{i=1}^{K} c_i(t)+r_K(t)}
\end{equation}
\begin{itemize}
  \item $C_{i}(t)$ indicate the $i^{\text{\tiny th}}$ extracted EM
  \item $r_{K}(t)$ indicate the residual
\end{itemize}

In this work, we computed HHS for each signal using the EMD to obtain a set of IMFs representing the original signal. Extracted features are Hilbert Spectrum (HS) and instantaneous energy density (IED) level. The decomposition into IMfs resulted in 10 IMFs per each channel.
\subsubsection{Band Power and RMS using DWT}
Discrete wavelet transform (DWT) is a recent technique of signal processing, it proceeds by the decomposition of the signal into different levels of approximation and detail corresponding to different frequency bands. It also keeps the temporal information of the signal. Compromise is done by downsampling the signal for each level. 

For example. Correspondence of frequency bands and wavelet decomposition levels depends on the sampling frequency. In our case, the correspondent decomposition is given in the last column of Table \ref{tab4} for $ f_s $= 128 Hz.
\begin{table}[!t]
\renewcommand{\arraystretch}{1.3}
\centering
\scriptsize
\caption{EEG signal Frequency Bands and Decomposition Levels at fs=128 Hz }
\label{tab4}
\begin{tabular}{c|c|c}
\hline \hline
\bfseries Bandwidth (Hz) & \bfseries Frequency Band & \bfseries Decomposition Level \\
\hline 
1-4 Hz & Delta $\delta$ & A5\\ \hline
4-8 Hz & Theta $\theta$& D5\\ \hline
8-13 Hz &Alpha $\alpha$ & D4\\ \hline
13-32 Hz & Beta $\beta$ & D3\\ \hline
32-64 Hz & Gamma $\Gamma$ & D2\\
\hline \hline
\end{tabular}
\end{table}
In our approach, in addition to the Band Power, the statistical feature Root Mean Square (RMS) derived from a wavelet decomposition with the function 'db5' for 5 levels, is extracted for each frequency band.

\begin{equation}
\footnotesize{
RMS(j)=\sqrt{\frac{\sum_{i=1}^{j}\sum_{n_{i}} D_i(n)^2}{\sum_{i=1}^{j}n_{i}}}}
\end{equation}
\normalsize
where $D_i$ are the detail coefficients, $n_i$ the number of $D_i$ at
the $i^th$ decomposition level, and j denotes the number of levels \cite{murugappan2010inferring}.

\subsection{Other Features}

\subsubsection{Quantitative Features}
In addition to the aforementioned features, we adopted the set of features used in \cite{toole2017neural}, which include a variation of commonly used EEG features.
However, for some features we use all channels and bands unlike in \cite{toole2017neural}  in which, a reduction of features was applied by averaging outputs.
The feature set includes stationary features that capture amplitude and frequency characteristics and inter-hemispheric connectivity features like shown in the Table \ref{tabqeeg}.\\
The column FB indicate if features are generated for each frequency band or not. 
\begin{table}[!t]
\renewcommand{\arraystretch}{1.3}
\centering
\scriptsize
\caption{EEG quantitative features }
\label{tabqeeg}
  \begin{tabular}{l|c}
  \hline \hline
    Feature Description & FB \\
    \hline 
    Absolute and relative(normalised to total spectral power) spectral power & Yes \\ \hline
    Spectral entropy: Wiener (measure of spectral flatness) & Yes \\ \hline
   Difference between consecutive short-time spectral estimates & Yes \\ \hline
    Cut-off frequency: 95\% of spectral power contained
    between 0.5 and fc Hz & No \\ \hline
    Amplitude: Time-domain signal: total power and standard deviation & Yes \\ \hline
    Amplitude: Skewness and of time-domain signal & Yes \\ \hline
    Amplitude: Envelope mean value and standard deviation (SD)& Yes  \\ \hline
    Connectivity: Brain Symmetry Index & Yes \\ \hline
    Connectivity: Correlation between envelopes of hemisphere-paired channels & Yes \\ \hline
    Connectivity: lag of maximum correlation coefficient  & \multirow{2}{*}{Yes} \\ 
    between hemisphere-paired channels & \\ \hline
    Connectivity: coherence: mean, maximum and frequency of maximum values & Yes \\ \hline
    Range EEG: mean, median, standard deviation and coefficient of deviation & Yes \\ \hline
    Range EEG: measure of skew about median & Yes \\ \hline
    Range EEG: lower margin (5th percentile) and upper margin (95th percentile) & Yes \\ \hline
    Range EEG: upper margin - lower margin & Yes \\ 
    \hline \hline
   \end{tabular}
\end{table}
\\
A feature vector resulting from this step containing a fusion of all qEEG features, constructed for ulterior classification.
\subsubsection{Differential Asymmetry}
Frontal asymmetry (the relative difference in power between two signals in different hemispheres) has been suggested as biomarker for anxiety \cite{demerdzieva2015relation}.
In different studies, FA was calculated from the beta band (13-25 Hz) or the gamma band ($>$ 30 Hz), but there are also studies using the alpha (8-12 Hz) frequency band. Due to the inverse relationship between alpha power and cortical activity, decreased alpha power reflects increased anxiety \cite{demerdzieva2015relation}.

Frontal asymmetry within the alpha band can be inversely related to stress/anxiety \cite{alarcao2017emotions} \cite{giannakakis2015detection}. The feature was calculated using alpha band power. The natural logarithm of left side channels were subtracted from the right ones (L-R).
\begin{equation}\label{}
\footnotesize{
  Asymmetry Index= \ln(\alpha)\bigg\rvert_{LChannel} - \ln(\alpha)\bigg\rvert_{RChannel}}
\end{equation}
To calculate the asymmetry index, the continuous signal must be broken into small parts. Scientific studies recommend overlapping epochs with each limited to a duration of 1-2 seconds \cite{zheng2017identifying}. For each epoch a DWT is calculated. The DWT determines which frequencies underlie the actual data, allowing you to extract the power in a specific frequency band.
\section{Classification}
In our Anxiety levels detection methodology, classification step is carried out with different classifiers which are SVM, k-NN and SSAE. 
\subsection{Support Vector Machines}
SVM \cite{vapnik1999overview} 
is known primarily as a binary classifier, while it can complete a multilabel classification. Known by its high generalization ability, it is tested and proven as a good classifier \cite{fourati2015improved}.
Consider a training set $(x_{j},y_{j}); 1\leq j\leq N$, where $x_{j}$ indicates the extracted EEG feature vectors, $y_{j}$
denotes the corresponding labels, and $N$ is the number of data.
The SVM decision function is expressed by the following equation:
\begin{equation}
\label{svm_eqn1}
f(x)=\sum_{i}^{N}\alpha_{i} y_{i}k(s_{i},x)+b
\end{equation}

$x$ denotes the the input vector extracted from the EEG signal, k is the kernel function, $s_{i}$  indicates support vectors, weights are denoted by $\alpha_{i}$ and the bias is denoted by b.
The kernel of SVM used in this work is the Radial Basis Function (RBF).

\subsection{k-Nearest Neighbor}
k-NN showed efficiency in EEG signal classification \cite{murugappan2010classification}. The k parameter is an integer constant always chosen par the user, where new case will be assigned to the class most common amongst its k nearest neighbours measured by an Euclidean distance, or Manhattan, Minkowski, and Hamming distances.
k-NN have an issue in classifying unbalanced sets, results can be influenced by the most dominant class. 

\subsection{Stacked Sparse Autoencoder}
Autoencoders seek to use items like feature selection and feature extraction to promote more efficient data coding. Autoencoders often use a technique called backpropagation to change weighted inputs, in order to achieve dimensionality reduction, which in a sense scales down the input for corresponding results. A sparse autoencoder is one that has small numbers of simultaneously active neural nodes.
The data processing in an auto-encoder network consists of two steps: 

Encoding, where the original data $x=[x^{(1)},x^{(2)},...,x^{(m)}]^{T}$  are encoded from the input layer to the hidden layer :
\begin{equation}
\label{ssae_eqn1}
y=f(x)=S(Wx+b)
\end{equation}
where $y=[y^{(1)},y^{(2)},...,y^{(n)}]^{T}$ indicates the feature expression of the hidden layer; $m$ is the number of nodes in the input layer. $n$ is the number of nodes in the output layer. $W$ denotes the weight matrix and $b$ design the bias vector. $S(x)$ is the sigmoid function, which is expressed as $S(x)=\frac{1}{(1+e^{-x})}$.

Decoding, the feature expression y is decoded from the hidden layer to the output layer:
\begin{equation}
\label{ssae_eqn2}
z = g(y) = S(W^{T}y + b^{'}) 
\end{equation}

The term Sparse AutoEncoder \cite{deng2013sparse} is used when sparse restrictions are added to the hidden noses of an autoencoder, in order to control the number of activated neurons. Sparse AutoEncoder shown better results in learning features regarding to its reduced number of activated neurons \cite{sun2016sparse}.

\section{Anxiety detection results and discussion}
\begin{figure}[!t]
  \centering
  \includegraphics[width=3.5in]{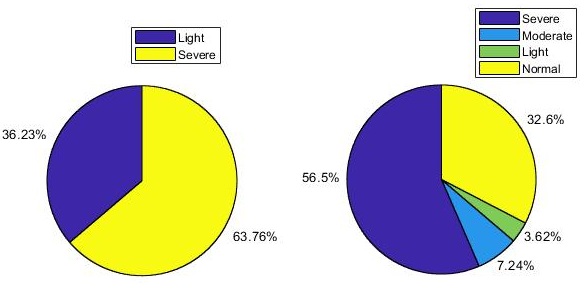}
  \caption{Overall class distribution across all participants for two and four anxiety levels.}\label{2classes}
\end{figure}
The main aim of the current work is to provide a new database for EEG-based anxiety detection. Therefore, to validate the proposed methodology three experiments basing on trial duration were conducted. As a preliminary step, data were labeled by applying an algorithm based on arousal and valence values to handle two classification problems which are anxiety two levels detection and anxiety four levels detection. The number of trials for each class resulting from this step is presented in Fig. \ref{2classes} in term of distribution percentage leading to unbalanced classes.

We believe that unbalanced data in the case of 4 anxiety levels affects the classification results. So, we propose in order to obtain a balanced data, to regroup classes two-by-two: normal and light in the first class and moderate and severe in the second class. The dataset becomes slightly unbalanced, with samples amounting to 36\% and 64\% in average for the first class (labeled light) and the second class (labeled severe) respectively.  Fig. \ref{2classes} shows the overall class distribution throughout the whole dataset for the two-class rating scale.

Classification was performed using a SVM classifier with Radial Basis Function (RBF) kernel. It was trained and tested in Matlab. Furthermore, a 5-fold cross validation technique was used in order to validate the classification performance. It must be noted that k-NN was also evaluated using the same procedure, and happened in some cases to produce a more significant results than SVM. We highlight that the task here is subject-independent that means the training and test samples does not belongs to the same subject.
 
 \begin{table}[!t]
\renewcommand{\arraystretch}{1.3}
 \centering
\caption{Anxiety detection results of 4 levels}
\label{tab4classes}
  \begin{tabular}{c|c|c|c|c}
  \hline \hline
\bfseries \multirow{2}{*}{Trial duration} & \bfseries \multirow{2}{*}{Feature} & \bfseries \multirow{2}{*}{\#Features} & \multicolumn{2}{c}{\textbf{Accuracy(\%)}}\\
\cline{4-5}
 & & & \bfseries SVM & \bfseries k-NN \\
    \hline
    \multirow{5}{*}{15s}&Hjorth&42&56.20&56.50 \\
   \cline{2-5}& qEEG& 25& 56.50&56.50 \\
   \cline{2-5}& HHT& 10& 57.00&56.80 \\
   \cline{2-5}& Power& 56&58.30&\textbf{57.60}\\
    \cline{2-5}& RMS& 56&\textbf{59.10}&56.50\\
  \hline
   \multirow{5}{*}{5s}&Hjorth&42&57.40&58.80\\
   \cline{2-5}& qEEG& 25& 56.80 &56.40\\
   \cline{2-5}& HHT& 9& 56.90 & 56.30\\
   \cline{2-5}& Power& 56 &62.00& 63.20\\
    \cline{2-5}& RMS& 56 &\textbf{65.30}& \textbf{64.30}\\
   \hline
    \multirow{5}{*}{1s}& Hjorth&42& 60.10&57.00 \\
   \cline{2-5}& qEEG& 25& 58.30 &56.40\\
   \cline{2-5}& HHT& 7& 56.60 &56.30 \\
   \cline{2-5}& Power& 56 &64.40 &68.00 \\
   \cline{2-5}& RMS& 56 &\textbf{70.20} & \textbf{73.60} \\
   \hline \hline
  \end{tabular}
  \end{table}

Anxiety detection results for 4 levels are presented in Table \ref{tab4classes}. We report results obtained from different kind of features. We remark that DWT-based RMS features with SVM achieve the best results 59.10\% and 65.30\% for 15s and 5s trial duration respectively. When the trial length is 1s, the result is increased with k-NN classifier again with DWT-based RMS features to reach 73.60\%. 
\begin{table}[!t]
\renewcommand{\arraystretch}{1.3}
\centering
\caption{Anxiety detection results of 2 levels}
\label{tab3}
  \begin{tabular}{c|c|c|c|c}
  \hline \hline
  \bfseries \multirow{2}{*}{Trial duration} & \bfseries \multirow{2}{*}{Feature} & \bfseries \multirow{2}{*}{\#Features} & \multicolumn{2}{c}{\textbf{Accuracy(\%)}}\\
\cline{4-5}
 & & & \bfseries SVM & \bfseries k-NN \\
 \hline
\multirow{4}{*}{15s}&Hjorth&42&66.30&63.80 \\  \cline{2-5}
  &qEEG&  25& 64.10 &63.80\\ \cline{2-5}
  & HHT&  10& 64.10 &64.10 \\ \cline{2-5}
  & Power&56& \textbf{66.30} &66.30 \\ \cline{2-5}
  & RMS&  56 & \textbf{66.30} & \textbf{67.00}\\ \cline{2-5}
  \hline
   \multirow{4}{*}{5s}&Hjorth&42&72.90&64.90 \\ \cline{2-5}
   & qEEG& 25& 64.00&63.60\\ \cline{2-5}
   & HHT& 9& 64.00& 64.10\\ \cline{2-5}
   & Power& 56 &\textbf{73.10}&70.50 \\ \cline{2-5}
   & RMS& 56 &72.90& \textbf{73.40}\\ \cline{2-5}
   \hline
    \multirow{4}{*}{1s}&Hjorth& 42& 67.40 & \textbf{81.40}\\\cline{2-5}
    & qEEG& 25& 64.00&63.50\\ \cline{2-5}
   & HHT& 7& 64.00& 63.60\\ \cline{2-5}
   & Power& 56  &76.00&74.90 \\ \cline{2-5}
   & RMS& 56 &\textbf{77.40} &80.30 \\ \cline{2-5}
   \hline \hline
  \end{tabular}
  \end{table}

Table \ref{tab3} presents the results of two anxiety levels (light and severe) detection. Power and RMS features obtained after a DWT for the 15 s EEG signals give a significant rates 66.30\%, 67.00\% with SVM and k-NN classifiers, respectively. 
For 5s Trial duration, 73.40\% is achieved with DWT based RMS features and k-NN classifier which slightly outperforms the result 73.10\% obtained with Power features and SVM classifier.
Classification accuracy reached 81.40\% using Hjorth parameters and k-NN classifier against 77.40\% with DWT based RMS features and SVM for 1s trial duration.

Through the aforementioned results, it is clear that detection from one second trial length is more accurate and this is related to the anxiety as an emotion. It can be evoked in 1s, but 5s or 15s are too long and may contain more than one emotion. To add, we can notice that the best rates are related to time-frequency features obtained after a wavelet transform. Regardless of the trials' duration, features produced from a Hilbert Hung Transform and the set of quantitative EEG features do not lead to a great accuracy, despite proving that this approach outperform rates in many researches. Knowing this, Hjorth parameters are the most simple features to extract from an EEG signal, yet they produce a significant accuracy throughout all types of datasets.
\begin{table}[!t]
\renewcommand{\arraystretch}{1.3}
\caption{Anxiety detection results using SSAE}
\label{SSAE4anx_table1}
\centering
\begin{tabular}{c|c|p{1.3cm}|c|c}
\cline{2-5}  \cline{2-5}
	& \bfseries \multirow{2}{*}{\#Features}  &\bfseries \multirow{2}{*}{L1-L2 sizes} & \multicolumn{2}{c}{\textbf{Accuracy(\%)}}\\
\cline{4-5} 
 & & & \bfseries 2 Levels & \bfseries 4 levels \\ 
 \hline
\multirow{2}{*}{\parbox{1.5cm}{Time Features}} & \multirow{2}{*}{67} &44-33 & \textbf{67.90} &57.80  \\ \cline{3-5}
							  &  &33-16 &67.60 & \textbf{59.90} \\   \hline \hline
\multirow{2}{*}{\parbox{1.5cm}{Frequency Features}} & \multirow{2}{*}{154} &102-77 & \textbf{82.00} & \textbf{71.20}  \\ \cline{3-5}
							  &  &77-38 &78.70 &68.80  \\   \hline
                              \hline
\multirow{4}{*}{\parbox{1.5cm}{Time-Frequency Features}} & \multirow{4}{*}{112 }& \multirow{2}{*}{74-56}  & \multirow{2}{*}{\textbf{67.10}}& \multirow{2}{*}{\textbf{59.70}}  \\      				&  &      &     &     \\ \cline{3-5}
                &  &\multirow{2}{*}{56-28} &\multirow{2}{*}{\textbf{67.10}}&\multirow{2}{*}{59.50}\\
                &  &      &     &      \\
                   \hline
                   \hline
\multirow{2}{*}{All Features}   & \multirow{2}{*}{277} &184-138  & \textbf{83.50} & \textbf{74.60}  \\ \cline{3-5}
                    &  &138-70 &81.60 &72.60 \\
                    \hline \hline
\end{tabular}
\end{table}
For further study the impact of feature type on the performance of the proposed system, Feature vectors from 1s trial are grouped by type and then passed to SSAE. As the role of SSAE is the selection of the most relevant features, we specified the size of both hidden layers to be lower than the input size. Note that, a softmax layer is added to perform the classification task. Table \ref{SSAE4anx_table1} depicts the results obtained for Time, Frequency, Time-Frequency and All Features. Frequency features outperform other types with 82\% and 71.20\% for 2 and 4 anxiety levels, respectively. The highest accuracy is obtained with the combination of all features to reach 83.50\% for 2 anxiety levels and 74.60\% for 4 anxiety levels. 

While, the combination of features allows to provide rich information, the representation generated by SSAE proved their effectiveness in handling more discriminative aspect. The result of 2 levels is higher than the 4 levels and this is mainly to the increase of complexity aspect in the classification task.

\section{Conclusion}
A novel approach to anxiety elicitation based on a face-to-face psychological stimulation has been presented in this paper. A dataset was constructed containing EEG data gathered during the experimentation. We present insights to analyze acquired data showing the efficiency of the followed strategy. The approach showed success in inducing anxiety levels, which was validated by HAM-A Test Scores calculated before and after the experiment. 

A various sets of emotion recognition features and in particular anxiety/stress, based on EEG signal, are reviewed and applied in this paper. We presented the most popular feature extraction techniques from the wide range used in the literature. Some methods perform slightly better than others. We also investigated which trial durations are most promising and which features are most effective for it. 

To the best of our knowledge, there are no available databases that contain EEG data recorded with a portable devices.
The headset Emotiv Epoc used in our work is available for everyone and is easy to install and use. Patients can use it at home and check their stress levels without consulting
an expert. Many clinical applications can be derived from this work improving life quality and reducing cognitive disabilities.


%

\section*{Acknowledgment}
The research leading to these results has received funding from the Ministry of Higher Education and Scientific Research of Tunisia under the grant agreement number LR11ES48.

\ifCLASSOPTIONcaptionsoff
  \newpage
\fi



%
\bibliographystyle{IEEEtran}
\bibliography{bibtex/IEEEabrv,bibtex/IEEEexample}{}
%

\begin{IEEEbiography}
[{\includegraphics[width=1in,height=1.25in,clip,keepaspectratio]{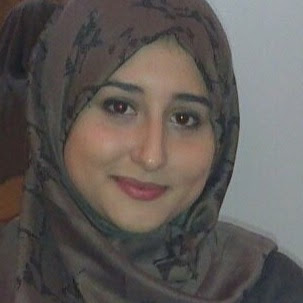}}]{Asma Baghdadi}(IEEE Graduated Student Member). A PhD student in Computer Systems Engineering at the National Engineering School of Sfax (ENIS) and at the University of Paris-Est. Graduated as Software engineer in 2014 from the Higher Institute of computer science of Tunis. She is currently a member of the REsearch Group in Intelligent Machines (REGIM). Also member of the Images, Signals and Intelligent Systems Laboratory  (Lissi-Upec). Since 2016 she is interested by researches in Affective computing, intelligent systems and EEG signal analysis.. 
\end{IEEEbiography}

\begin{IEEEbiography}
[{\includegraphics[width=1in,height=1.25in,clip,keepaspectratio]{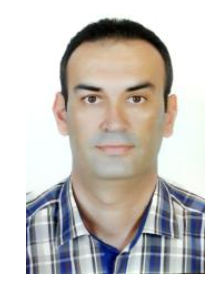}}]{Yassine Aribi}
(IEEE Student Member'08, Senior Member'16). He received the B. S. Degree in Computer Science applied in the management from the Faculty of Economic Sciences and Management of Sfax (FSEG) in 2002, the M. S. in Novel Technologies for Computer Systems from the National Engineering School of Sfax (ENIS) in 2011 and the PhD degree in Computer Science from the National Engineering School of Sfax (ENIS), Tunisia in 2015. In 2003, he joined Gab\`es University, where he is a common trunk professor in the Department of Computer Science, and a research member in the Research Groups on Intelligent Machines (REGIM) from 2010. In 2015, he joined the University of Monastir, where he is currently an assistant at the Higher Institute of computer science of Mahdia(Tunisia).  His research include Medical images processing using intelligent tools. 
\end{IEEEbiography}

\begin{IEEEbiography}
[{\includegraphics[width=1in,height=1.25in,clip,keepaspectratio]{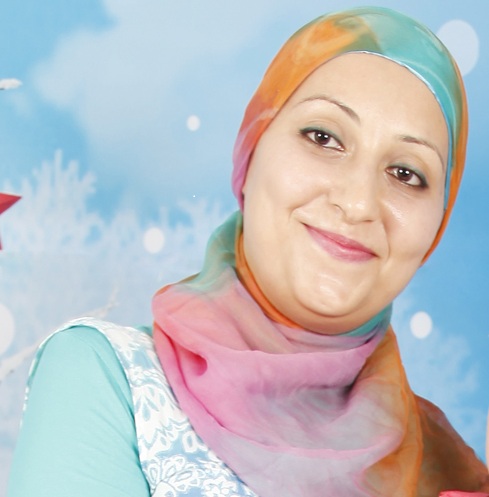}}]{Rahma Fourati}
(IEEE Graduated Student Member'15). She was born in Sfax, she is a PhD student in Computer Systems Engineering at the National Engineering School of Sfax (ENIS), since October 2015. She received the M.S. degree in 2011 from the Faculty of Economic Sciences and Management of Sfax (FSEGS). She is currently a member of the REsearch Group in Intelligent Machines (REGIM).Her research interests include Recurrent Neural Networks, Affective computing, EEG signals analysis, Support Vector Machines, Simulink Modeling. 
\end{IEEEbiography}

\begin{IEEEbiography}
[{\includegraphics[width=1in,height=1.25in,clip,keepaspectratio]{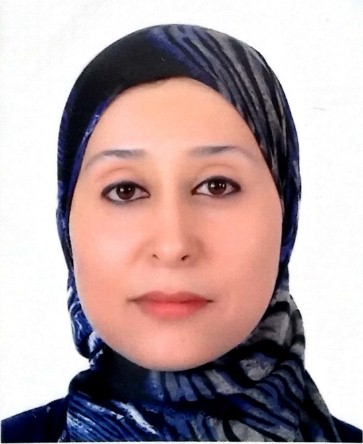}}]{Najla Halouani}
Najla Halouani : She was born in Sfax  She is an Assistant University Hospital in Psychiatry since October 2012. 
She received the Doctorate of State in Medicine in 2011 from the Faculty of Medicine of Sfax (FMS). She received the Simulation Diploma in Medical Sciences in 2014. Her research interests include Diagnosis and therapy of stress and anxiety, Depression Study, Post-traumatic Stress Disorder.
\end{IEEEbiography}

\begin{IEEEbiography}
[{\includegraphics[width=1in,height=1.25in,clip,keepaspectratio]{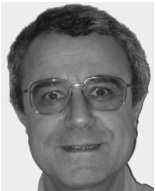}}]{Patrick Siarry}
(SM'95) is currently a Professor and
a Doctoral Supervisor of Automatics and Informatics
with  the   University   of  Paris-Est  Cr{\'e}teil,   Cr{\'e}teil, France.  He  is  the  Head  of  one  team  of  the  Laboratoire  Images,  Signaux  et  Syst{\`e}mes  Intelligents, University   of   Paris-Est   Cr{\'e}teil.   He   supervises   a Research  Working  Group,  in  the  frame  of  Centre National de la Recherche Scientifique, META, since 2004. META is interested in theoretical and practical advances  in  metaheuristics   for  hard  optimization.
His  current  research  interests  include  improvement of existing  metaheuristics  for hard optimization,  adaptation  of discrete  meta-heuristics  to  continuous  optimization,
hybridization  of  metaheuristics  with classical  optimization  techniques,  dynamic  optimization,  and  particle  swarm optimization.
\end{IEEEbiography}
\begin{IEEEbiography}
[{\includegraphics[width=1in,height=1.25in,clip,keepaspectratio]{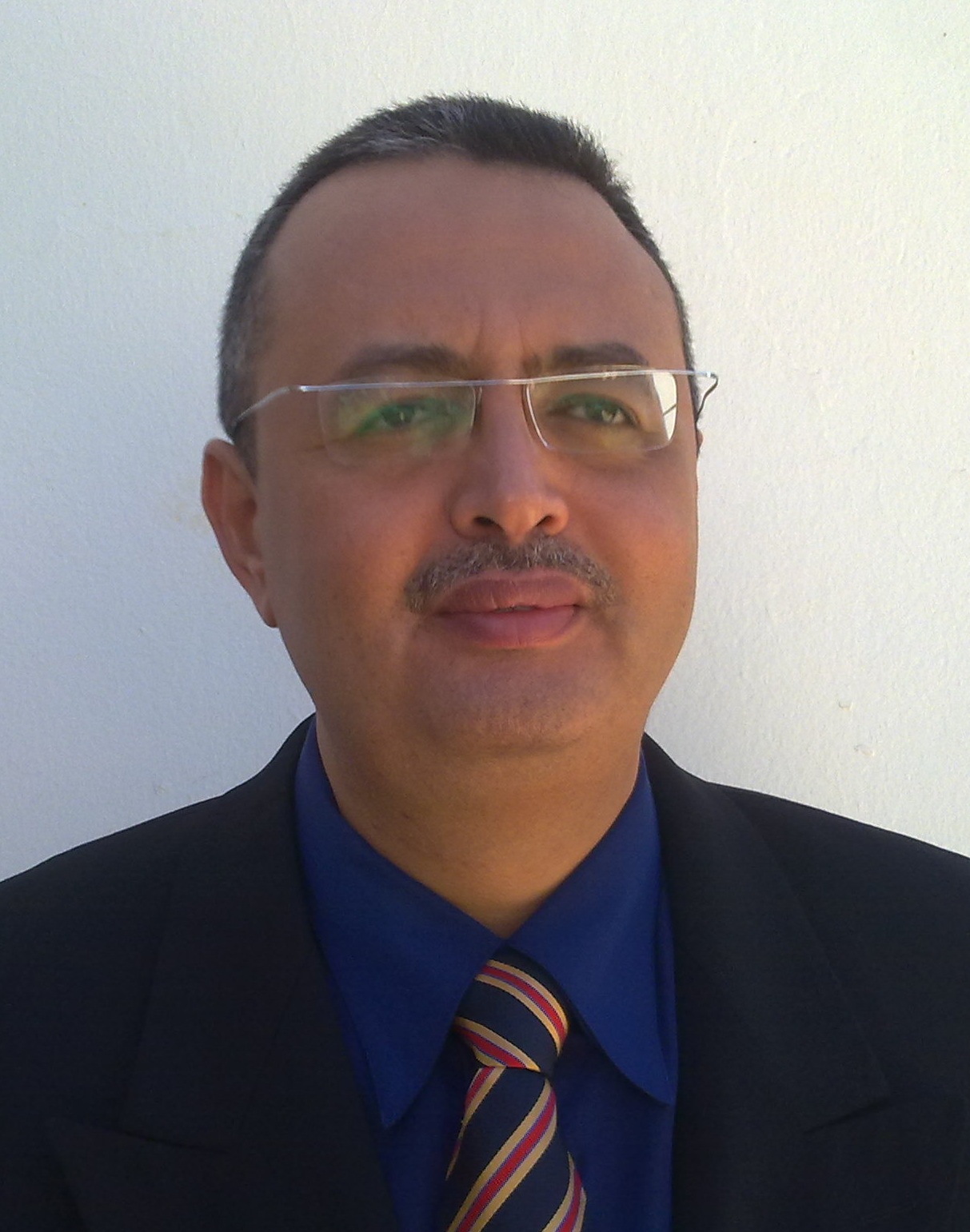}}]{Adel M. Alimi}
(IEEE Student Member'91, Member'96, Senior Member'00).He was born in Sfax (Tunisia) in 1966. He graduated in Electrical Engineering 1990, obtained a Ph.D. and then an HDR both in Electrical and Computer Engineering in 1995 and 2000 respectively. He is now a Professor in Electrical and Computer Engineering at the University of Sfax. His research interest includes applications of intelligent methods (neural networks, fuzzy logic, evolutionary algorithms) to pattern recognition, robotic systems, vision systems, and industrial processes. He focuses his research on intelligent pattern recognition, learning, analysis and intelligent control of large scale complex systems. He is an Associate Editor and Member of the editorial board of many international scientific journals (e.g. Pattern Recognition Letters, Neurocomputing, Neural Processing Letters, International Journal of Image and Graphics,
Neural Computing and Applications, International Journal of Robotics and Automation, International Journal of Systems Science, etc.). He was a Guest Editor of several special issues of international journals (e.g. Fuzzy Sets and Systems, Soft Computing, Journal of Decision Systems, Integrated Computer Aided Engineering, Systems Analysis Modeling and Simulations). He is an IEEE senior member and member of IAPR, INNS and PRS
\end{IEEEbiography}





\end{document}